\documentclass[letterpaper]{article} 
\usepackage{aaai2027} 
\usepackage[hyphens]{url} 
\usepackage{graphicx} 
\usepackage{natbib} 
\usepackage{caption} 
\usepackage{booktabs}
\usepackage{multirow}
\usepackage{colortbl}
\usepackage{amsmath}
\usepackage{amssymb}

\definecolor{tracehighlight}{gray}{0.92}

\title{TRACE-Memory: Public-Conditioned Retrieval and Utility-Aware Evidence Admission for Personalized Generation}
\author{
Jing Wang, Zhu Wang\corresponding, Yifan Guo, Yulong Yang, Yunji Liang
}
\affiliations{
Northwestern Polytechnical University\\
jingwang25@mail.nwpu.edu.cn, wangzhu@nwpu.edu.cn
}

\begin{document}

\maketitle

\begin{abstract}
Personalized generation systems retrieve user history by request--memory
relevance and inject it into the model context. Yet relevant history may
concern the wrong preference aspect, duplicate public information, or provide
insufficient support. We argue that personal memory should be used only when
it adds utility beyond a public-only response. We propose TRACE-Memory, a
two-stage framework for selective personalization. Stage 1 queries for
user-specific information missing from the request and public context, then
retrieves a coverage-oriented candidate pool. Stage 2 admits a compact subset
of source-traceable evidence units, or the empty set, according to
response-level incremental utility. We progressively train the query-generation
and evidence-admission policies through structured SFT initialization,
reduced-space stage-wise GRPO warm-up, and nested multi-sample Joint GRPO.
On a 5,400-task benchmark spanning Controlled and Natural tasks from Goodreads,
Amazon Reviews, and Reddit, TRACE-Memory consistently outperforms random and lexical memory use,
improves over semantic retrieval, remains competitive with frontier-LLM memory
pipelines as local generator capacity increases, and conditions evidence
admission on public-context sufficiency, supporting selective rather than
default personalization.
\end{abstract}


\section{Introduction}

Large language models can produce informative responses from a current
request, public information about the target, and general knowledge encoded in
their parameters. This public-only path, however, does not directly represent
how a particular user evaluates an attribute, balances competing factors, or
reacts to similar items. Historical reviews, choices, and interactions provide
user-specific evidence beyond public information, making personal-history
retrieval central to personalized generation
\citep{salemi2024lamp,mysore2024pearl,salemi2024retrieveroptimization}.

\begin{figure}[t]
  \centering
  \includegraphics[width=\columnwidth]{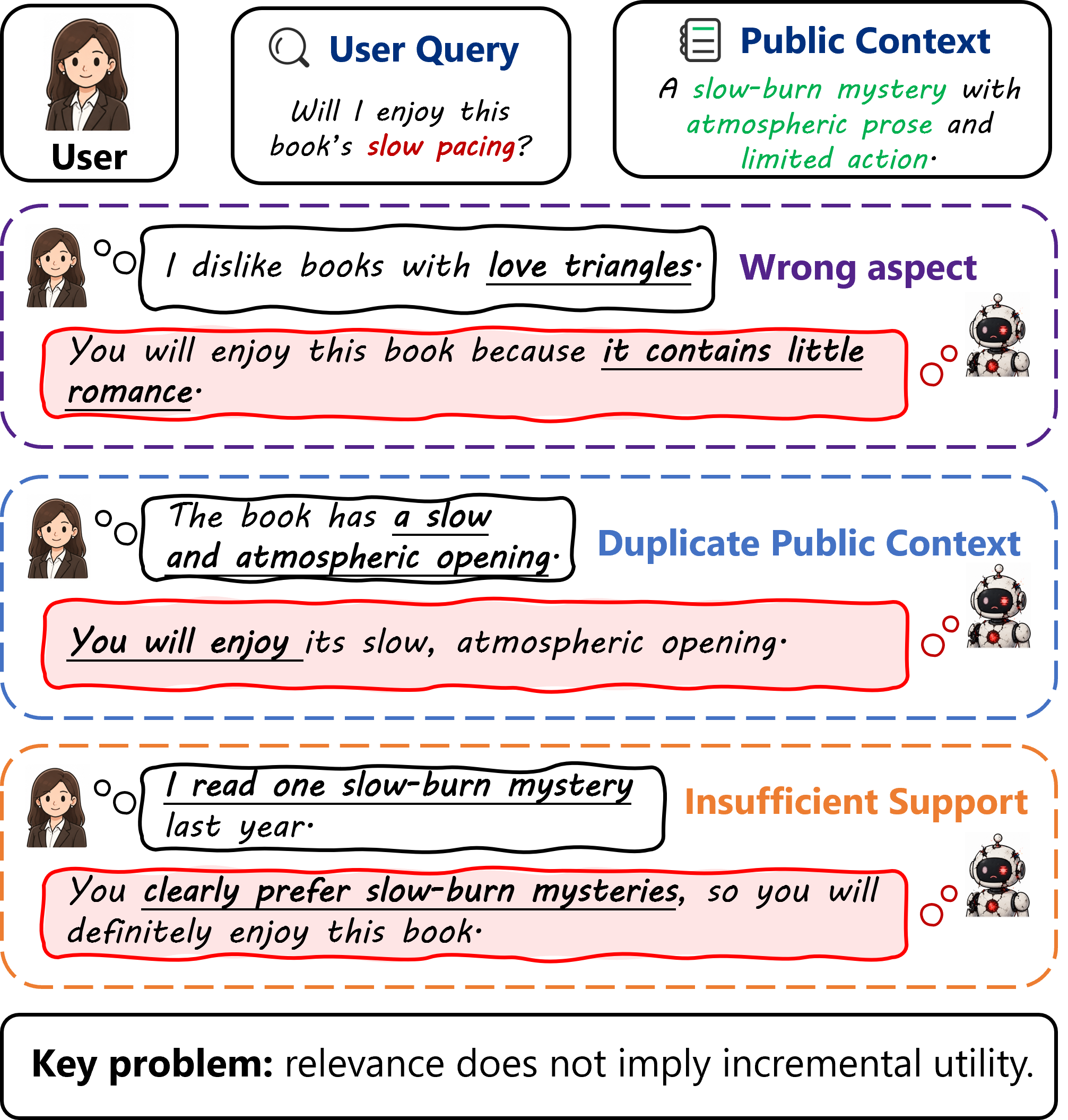}
  \caption{Three failure modes of relevance-based memory use.}
  \label{fig:memory-failure-modes}
\end{figure}

Yet relevance to the request alone does not warrant placing a memory in the
generation context. Request-level similarity may retrieve history about
the wrong preference aspect; a relevant record may provide too little evidence
for a stable user judgment; and even accurate personal information may merely
repeat what the public context already provides. Injecting such history
increases context cost and can produce irrelevant, unsupported, or unnecessary
personalization. Semantic relevance therefore does not imply incremental
utility beyond the public-only response path
\citep{hu2026opbench,liu2024lost,yoran2024robust,shi2023distracted}.
Figure~\ref{fig:memory-failure-modes} illustrates these three failure modes.

Existing personalized-memory systems commonly rank user histories using
lexical or semantic similarity, retrieval recall, or LLM-based selection
before injecting the resulting context
\citep{mysore2024pearl,salemi2024retrieveroptimization,xu2024recomp}.
They primarily ask which records match the request, rather than what
user-specific information remains missing after the request and public context
are given, or whether the retrieved evidence will improve the final response.
A selective system must therefore identify the missing preference aspect,
reject weak evidence, and abstain when the public context is already
sufficient. Although response-aware selection has begun to connect memory use
with downstream model behavior~\citep{fisher2026rums}, selective
personalization still requires a public-conditioned information-gap model and
an explicit option to use no personal evidence.

To address this issue, we propose \textbf{TRACE-Memory} (\textbf{T}ask-conditioned
\textbf{R}etrieval and \textbf{A}dmission of \textbf{C}ontextual
\textbf{E}vidence), a two-stage framework for conditional personal-evidence
governance. Stage 1 generates public-conditioned queries that describe missing
user preferences, reactions, or behavioral patterns, and uses them to build a
coverage-oriented candidate pool. The retrieved memories are segmented into
source-traceable evidence units. Stage 2 then admits a compact subset according
to its response-level incremental utility relative to the public-only path.
An explicit empty-set action lets the system abstain from personalization when
history adds insufficient value.

The policies are learned through a progressive curriculum comprising
structured SFT, reduced-space stage-wise GRPO, and nested multi-sample Joint
GRPO. Nested reward aggregation connects downstream evidence utility to
upstream query generation without requiring simultaneous policy updates. We
construct a 5,400-task benchmark spanning Controlled and Natural tasks from
Goodreads, Amazon Reviews, and Reddit, and evaluate multiple policy backbones
and response generators.
Experiments reveal three consistent patterns. TRACE-Memory improves over
random and lexical memory use in all six fixed-generator settings and over
semantic retrieval in four, confirming that response utility is a stronger
admission signal than relevance alone. As local generator capacity increases,
the average gap to frontier-LLM pipelines shrinks from $-0.28/-0.29$ at 8B to
$0.00$ at 32B, showing that evidence governance can recover a substantial part
of the system-level capability gap. Finally, from L0 to L3, TRACE-Memory raises
its EMPTY rate from 7.69\% to 12.53\% while average reward increases from 0.31
to 0.62, demonstrating context-sensitive rather than default personalization.

\noindent\textbf{Contributions.}
Our main contributions are:
\begin{itemize}
  \item We formulate personal-memory use as incremental evidence utility over
  a public-only response path, separating useful personalization from semantic
  relevance.

  \item We propose TRACE-Memory with public-conditioned, coverage-oriented
  candidate construction, utility-aware evidence admission, and explicit
  empty-set support.

  \item We develop a progressive policy-optimization curriculum with nested
  reward aggregation, and evaluate its candidate coverage, evidence quality,
  selective memory use, system-level competitiveness, and final response
  quality across multiple domains and model scales.
\end{itemize}

\section{Related Work}

\subsection{Personalization with User Memory}

Personalized generation has evolved from conditioning on static personas or
predefined user attributes~\citep{zhang2018persona,zheng2019personalizeddialogue}
to using longitudinal histories at inference time. LaMP and LongLaMP establish
retrieval-augmented benchmarks across short- and long-form personalized tasks
\citep{salemi2024lamp,kumar2024longlamp}, while PEARL and ROPG optimize history
retrieval with downstream generation feedback
\citep{mysore2024pearl,salemi2024retrieveroptimization}. Parameter-efficient
approaches instead encode preferences through shared adapters, user
representations, or plug-in modules
\citep{tan2024personalizedpieces,liu2025personaplug}.

Long-term memory systems and benchmarks further study persistent experience,
long-horizon recall, preference following, dynamic profiling, and incremental
memory management
\citep{park2023generativeagents,maharana2024locomo,zhao2025prefeval,
jiang2025personamem,hu2025memoryagentbench}. These works establish the value of
preserving and retrieving user history, but recall alone does not determine
whether a memory should enter the current response. TRACE-Memory assumes an
existing history and instead studies conditional evidence admission given the
request and public context.

\begin{figure*}[t]
  \centering
  \includegraphics[width=0.98\textwidth]{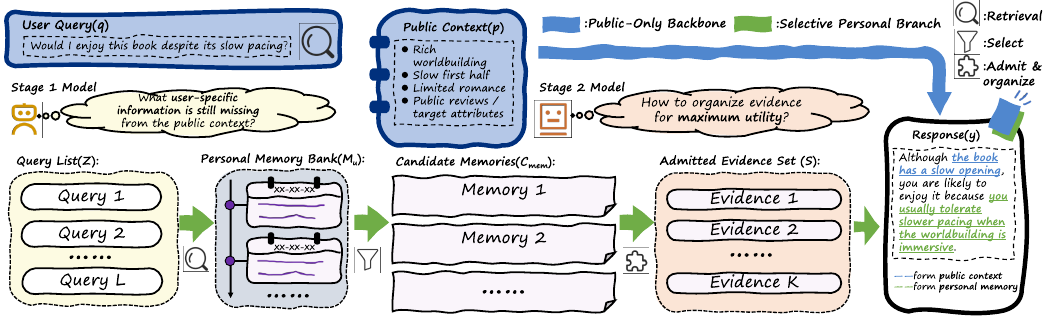}
  \caption{Overview of TRACE-Memory. Stage 1 generates public-conditioned
  memory queries to retrieve candidate memories, and Stage 2 admits a compact
  evidence set for personalized response generation.}
  \label{fig:trace-overview}
\end{figure*}

\subsection{Adaptive Retrieval and Selective Personalization}

RAG and REALM established retrieval as a non-parametric knowledge source
\citep{lewis2020retrieval,guu2020realm}. Later methods make retrieval and
augmentation conditional on uncertainty, task complexity, evidence quality,
or estimated usefulness
\citep{jiang2023flare,jeong2024adaptiverag,asai2024selfrag,xu2024recomp}.
Related studies show that irrelevant context can distract language models,
supporting selective filtering rather than fixed context injection
\citep{yoran2024robust,shi2023distracted}.

This issue is sharper for personal memory, where unnecessary history can cause
irrelevance, repetition, or unsupported user inference. OP-Bench directly
studies such over-personalization~\citep{hu2026opbench}, while selective
prediction provides the broader principle that a model may abstain when
evidence is insufficient~\citep{geifman2017selective}. Unlike adaptive RAG,
which mainly asks whether external knowledge is needed, TRACE-Memory treats the
empty evidence set as the exact public-only path and admits personal evidence
only for its incremental value over that path.

\subsection{Response-Aware Memory Selection}

LLM-generated query expansion and rewriting can improve retrieval beyond the
original request~\citep{wang2023query2doc,gao2023hyde,ma2023queryrewrite}, but
these methods do not model which user information remains missing after public
context is observed. More closely related work uses downstream generation
signals to calibrate retrieval, including REPLUG, PEARL, and ROPG
\citep{shi2024replug,mysore2024pearl,salemi2024retrieveroptimization}. Their
primary decision remains document ranking for a given query.

RUMS is the closest response-aware memory-selection method: it estimates memory
utility from its information-theoretic effect on the response distribution and
can return an empty set~\citep{fisher2026rums}. TRACE-Memory differs in three
ways. Utility is measured relative to a response already conditioned on public
context; candidate coverage is separated from final admission over
source-traceable evidence~\citep{gao2023alce}; and downstream response utility
is propagated to public-conditioned query generation. The resulting problem is
conditional evidence governance rather than static retrieval or independent
memory-subset selection.

\section{Method}

\subsection{Problem Definition}

We consider a personalized generation task with a user request $q$, public
context $p$, and user memory bank
$\mathcal{M}_u=\{m_{u,1},\ldots,m_{u,N}\}$. The public context contains
task-level information available independently of the target user, such as
item attributes, public reviews, and population-level signals, but excludes
general knowledge encoded in the generator.

Let $G$ be the frozen response generator. A deterministic compiler $\Gamma(X)$
orders and formats selected personal evidence $X$ for $G$, without retrieval,
selection, or semantic rewriting. We set $\Gamma(\varnothing)=\epsilon$, making
the empty evidence set exactly the public-only generation path. A
relevance-based system instead retrieves
$C=\mathcal{R}(q,p,\mathcal{M}_u)$ and generates
$y\sim p_G(\cdot\mid q,p,\Gamma(C))$, although relevance does not establish
value beyond $p$ or improvement in the final response.

Ideally, a selected personal-evidence set $S$ is valued by its utility gain over
the public-only path:
\begin{equation}
\begin{split}
\Delta U_u(S;q,p)
={}&
\mathbb{E}_{y\sim p_G(\cdot\mid q,p,\Gamma(S))}
\left[U_u(y)\right]\\
&-
\mathbb{E}_{y\sim p_G(\cdot\mid q,p)}
\left[U_u(y)\right].
\end{split}
\end{equation}
Because true utility $U_u$ is unobserved during training, TRACE-Memory
approximates it through coverage-oriented candidate construction and
utility-aware evidence admission.

\subsection{TRACE-Memory Overview}

Figure~\ref{fig:trace-overview} shows the full pipeline. Stage 1 maps $(q,p)$
to public-conditioned queries $Z$ that describe missing user-specific
information. A frozen retriever and deterministic segmenter construct
source-traceable candidate evidence, after which Stage 2 selects a compact set
$S$ or returns the empty set. The two structured policies are initialized by
SFT and progressively optimized over the nested query--evidence action space;
retrieval, segmentation, context compilation, and response generation remain
frozen.

\subsection{Public-Conditioned Query Generation}

Directly scoring a memory's incremental value before retrieval is difficult
because that value depends jointly on the task and what the public context
already provides. Stage 1 instead predicts the user-specific information
missing from $p$ and expresses these gaps as retrieval queries, turning
unobserved value estimation into coverage-oriented retrieval.

\begin{figure*}[t]
  \centering
  \includegraphics[width=0.98\textwidth]{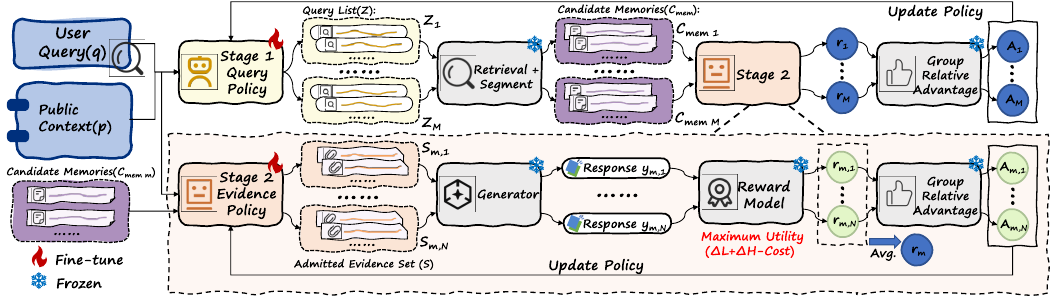}
  \caption{Nested multi-sample Joint GRPO in TRACE-Memory. Stage 1 samples
  multiple query lists, and Stage 2 samples multiple evidence actions under
  each query-induced candidate pool. Each Stage 2 action receives direct
  evidence utility, while the mean utility within its action group defines the
  reward of the corresponding Stage 1 query. Flame and snowflake icons denote
  trainable and frozen modules, respectively.}
  \label{fig:trace-training}
\end{figure*}

Unlike generic query expansion and rewriting
\citep{wang2023query2doc,gao2023hyde,ma2023queryrewrite}, TRACE-Memory uses a
query policy $\pi_{\phi}$ to generate a canonical list of
public-conditioned information gaps:
\begin{equation}
\begin{split}
Z
={}&
(z_1,z_2,\ldots,z_L),\\
Z
\sim{}&
\pi_\phi(\cdot\mid q,p),
\qquad
L\leq L_{\max}.
\end{split}
\end{equation}
Each $z_i$ describes a missing user preference, reaction, tolerance, or
behavioral pattern rather than repeating public facts about the target item.

The retriever executes these queries over personal history and aggregates
their results under a shared budget:
\begin{equation}
\begin{split}
C_{\mathrm{mem}}
={}&
\mathcal{R}(Z,\mathcal{M}_u)\\
={}&
\operatorname{Aggregate}\!\left(
\bigcup_{z_i\in Z}
\operatorname{Retrieve}(z_i,\mathcal{M}_u)
\right).
\end{split}
\label{eq:candidate-memory-set}
\end{equation}
Retrieval uses BGE-base-en-v1.5 embeddings~\citep{xiao2023cpack}. Stage 1 is
deliberately recall-oriented: it maximizes the opportunity to recover useful
personal information but leaves admission to Stage 2.

\subsection{Utility-Aware Evidence Admission}

The coverage-oriented pool may contain redundant, weakly supported, or
publicly covered memories. Stage 2 converts it into a precision-oriented set by
evaluating the incremental utility of the evidence that would reach the
generator.

\noindent\textbf{Evidence action.}
$\Psi$ produces fine-grained evidence units:
\begin{equation}
C_{\mathrm{evi}}
=
\Psi(C_{\mathrm{mem}})
=
\{e_1,e_2,\ldots,e_K\}.
\end{equation}
Each unit retains its source identifier, offsets, and timestamp, keeping
selected evidence attributable to its original interaction
\citep{gao2023alce}. Stage 2 then outputs:
\begin{equation}
\begin{split}
A
={}&
\bigl(
\langle id_1,a_1\rangle,\ldots,
\langle id_J,a_J\rangle,\texttt{STOP}
\bigr)\\
\sim{}&
\pi_\omega(\cdot\mid q,p,Z,C_{\mathrm{evi}}),
\qquad
S(A)\subseteq C_{\mathrm{evi}},
\end{split}
\end{equation}
where $a_j$ is an extractive anchor. Conditioning on $Z$ preserves the
information-gap intent, while $\texttt{EMPTY}$ maps to
$S(A)=\varnothing$.

\noindent\textbf{Incremental utility.}
Stage 1 targets potential value; Stage 2 evaluates realized utility. Because
$U_u$ is unavailable, we measure how $S$ changes the frozen generator's
distribution along a target response $y^*$ relative to the public-only path.
The target is used only for offline rewards, never as input to either policy or
the retriever. At position $t$, let
\begin{equation}
\begin{aligned}
p_t^S
&=p_G(\cdot\mid y_{<t}^*,q,p,\Gamma(S)),\\
p_t^\varnothing
&=p_G(\cdot\mid y_{<t}^*,q,p).
\end{aligned}
\end{equation}
We use per-token likelihood gain and entropy reduction:
\begin{equation}
\begin{aligned}
\Delta L(S)
&=T^{-1}\sum_{t=1}^{T}
\left[\log p_t^S(y_t^*)-\log p_t^\varnothing(y_t^*)\right],\\
\Delta H(S)
&=T^{-1}\sum_{t=1}^{T}
\left[H(p_t^\varnothing)-H(p_t^S)\right].
\end{aligned}
\label{eq:entropy-reduction}
\end{equation}
Positive $\Delta L(S)$ means that the evidence better explains the observed
response; positive $\Delta H(S)$ means that it reduces uncertainty along that
path.

The evidence reward combines these utility signals with compactness and
validity constraints:
\begin{equation}
\begin{aligned}
R_E(A)
={}&\Delta L(S(A))\\
&+\eta\,\mathbf{1}\!\left[
\Delta L(S(A))\geq-\epsilon_L
\right]\Delta H(S(A))\\
&-\lambda_E\operatorname{Cost}_E(S(A))
-\lambda_V\mathbf{1}\!\left[\operatorname{Invalid}(A)\right],\\
\operatorname{Cost}_E(S)
={}&\alpha_E|S|+\beta_E\operatorname{Len}(\Gamma(S)).
\end{aligned}
\label{eq:evidence-utility}
\end{equation}
Invalid actions include malformed outputs, unavailable evidence identifiers,
and ID--anchor inconsistencies. Since $\Gamma(\varnothing)=\epsilon$,
$R_E(\texttt{EMPTY})=0$.

\subsection{Progressive Warm Start and Joint GRPO}

Training proceeds from SFT through reduced-space stage-wise GRPO to nested
multi-sample Joint GRPO. This curriculum limits early branching before full
query--evidence exploration. Both policies use group-relative policy
optimization (GRPO)~\citep{shao2024deepseekmath}.

\noindent\textbf{Structured SFT initialization.}
The two policies are first trained independently on teacher-provided
public-conditioned queries and source-grounded evidence actions:
\begin{equation}
\begin{aligned}
\mathcal{L}_{\mathrm{SFT}}^{(1)}
&=
-\mathbb{E}
\left[
\log \pi_{\phi}(Z^*\mid q,p)
\right],\\
\mathcal{L}_{\mathrm{SFT}}^{(2)}
&=
-\mathbb{E}
\left[
\log \pi_{\omega}(A^*\mid q,p,Z,C_{\mathrm{evi}})
\right].
\end{aligned}
\end{equation}
SFT ensures valid actions but does not optimize utility.

\noindent\textbf{Reduced-space stage-wise warm-up.}
We freeze SFT-initialized Stage 1 and train Stage 2 in one query-induced
candidate space per task, then freeze Stage 2 and train Stage 1 with one
downstream evidence rollout per query. This restricted branching warms up
utility-oriented behavior before joint exploration; short SFT refreshes
preserve valid outputs.

\noindent\textbf{Nested multi-sample Joint GRPO.}
After warm-up, we expand to the nested action space in
Figure~\ref{fig:trace-training}. For each task
$(q,p,\mathcal{M}_u,y^*)$, Stage 1 samples $M$ query lists and Stage 2 samples
$N$ evidence actions per induced pool:
\begin{equation}
\begin{aligned}
Z_m
&\sim
\pi_\phi(\cdot\mid q,p),
&&m=1,\ldots,M,\\
A_{m,n}
&\sim
\pi_\omega(\cdot\mid q,p,Z_m,C_{\mathrm{evi},m}),
&&n=1,\ldots,N.
\end{aligned}
\end{equation}
From each $Z_m$, retrieval yields $C_{\mathrm{mem},m}$ and segmentation yields
$C_{\mathrm{evi},m}$.

Each Stage 2 action receives its direct evidence reward. The mean reward of the
$N$ Stage 2 actions under one query becomes the sample-level reward of that
Stage 1 query:
\begin{equation}
\begin{aligned}
&r_{m,n}^{(2)}=R_E(A_{m,n}),\quad
\bar r_m^{(2)}=N^{-1}\sum_{n=1}^{N}r_{m,n}^{(2)},\\
&r_m^{(1)}=\bar r_m^{(2)}
-\lambda_Q\operatorname{Cost}_Q(Z_m),\\
&\operatorname{Cost}_Q(Z_m)
=\alpha_Q|Z_m|+\beta_Q\operatorname{Len}(Z_m).
\end{aligned}
\label{eq:nested-stage-rewards}
\end{equation}
The within-query mean $\bar r_m^{(2)}$ is one Stage 1 sample reward, not its
GRPO group mean.

\noindent\textbf{Group-relative updates.}
Stage 2 normalizes rewards across the $N$ evidence actions conditioned on the
same query $Z_m$. Stage 1 separately normalizes the resulting query rewards
across the $M$ query lists sampled for the same task:
\begin{equation}
\begin{aligned}
&A_{m,n}^{(2)}
=\frac{r_{m,n}^{(2)}-\bar r_m^{(2)}}
{\sigma_m^{(2)}+\epsilon_A},\quad
\bar r^{(1)}=M^{-1}\sum_{m=1}^{M}r_m^{(1)},\\
&A_m^{(1)}
=\frac{r_m^{(1)}-\bar r^{(1)}}
{\sigma^{(1)}+\epsilon_A}.
\end{aligned}
\end{equation}
Each structured action receives one group-relative advantage. Both policies
use standard GRPO with token-level clipping and KL regularization to their SFT
references, following the clipped-update principle of
PPO~\citep{schulman2017ppo}. The other policy remains frozen during each update.

Joint GRPO explores and scores nested $M\times N$ query--evidence actions; the
policies are updated separately.

\noindent\textbf{Inference.}
At test time, both policies decode deterministically without a target response:
\begin{equation}
\begin{aligned}
&\hat Z=\operatorname{Decode}\!\left(\pi_\phi(\cdot\mid q,p)\right),\\
&\hat C_{\mathrm{mem}}=\mathcal{R}(\hat Z,\mathcal{M}_u),\quad
\hat C_{\mathrm{evi}}=\Psi(\hat C_{\mathrm{mem}}),\\
&\hat A=\operatorname{Decode}\!\left(
\pi_\omega(\cdot\mid q,p,\hat Z,\hat C_{\mathrm{evi}})
\right),\\
&\hat S=S(\hat A),\quad
y\sim p_G(\cdot\mid q,p,\Gamma(\hat S)).
\end{aligned}
\label{eq:evidence-selection}
\end{equation}

\section{Experiments}

\begin{table*}[t]
  \centering
  {\small
    \setlength{\tabcolsep}{3pt}
    \begin{tabular*}{\textwidth}{@{\extracolsep{\fill}}lcccccc@{}}
      \toprule
      & \multicolumn{3}{c}{Controlled} & \multicolumn{3}{c}{Natural} \\
      \cmidrule(lr){2-4}\cmidrule(lr){5-7}
      Baseline
      & \shortstack{DeepSeek-\\V4-Flash}
      & \shortstack{Gemini 3.1\\Flash-Lite}
      & GPT-5.4
      & \shortstack{DeepSeek-\\V4-Flash}
      & \shortstack{Gemini 3.1\\Flash-Lite}
      & GPT-5.4 \\
      \midrule
      Public-only
        & \cellcolor{tracehighlight}\textbf{0.12} & \cellcolor{tracehighlight}\textbf{0.18}
        & \cellcolor{tracehighlight}0.04 & -0.02
        & \cellcolor{tracehighlight}0.03 & -0.05 \\
      Random
        & \cellcolor{tracehighlight}\textbf{0.29} & \cellcolor{tracehighlight}\textbf{0.38}
        & \cellcolor{tracehighlight}\textbf{0.15} & \cellcolor{tracehighlight}\textbf{0.23}
        & \cellcolor{tracehighlight}\textbf{0.11} & \cellcolor{tracehighlight}\textbf{0.14} \\
      All
        & \cellcolor{tracehighlight}\textbf{0.22} & \cellcolor{tracehighlight}\textbf{0.15}
        & -0.08 & \cellcolor{tracehighlight}0.09 & -0.02 & \textbf{-0.15} \\
      BM25
        & \cellcolor{tracehighlight}0.05 & \cellcolor{tracehighlight}\textbf{0.14}
        & \cellcolor{tracehighlight}\textbf{0.17} & \cellcolor{tracehighlight}0.10
        & \cellcolor{tracehighlight}0.07 & \cellcolor{tracehighlight}0.07 \\
      Semantic similarity
        & \cellcolor{tracehighlight}\textbf{0.12} & \cellcolor{tracehighlight}\textbf{0.11}
        & -0.04 & \cellcolor{tracehighlight}0.03
        & \cellcolor{tracehighlight}0.09 & -0.01 \\
      LLM+LLM
        & -0.02 & \textbf{-0.11} & \textbf{-0.12} & -0.06 & -0.07
        & \cellcolor{tracehighlight}0.06 \\
      Rule-based+Semantic
        & \cellcolor{tracehighlight}0.06 & -0.01 & -0.04
        & \cellcolor{tracehighlight}0.03 & \cellcolor{tracehighlight}0.08
        & \cellcolor{tracehighlight}0.02 \\
      \bottomrule
    \end{tabular*}
  }
  \caption{Signed preference advantage of TRACE-Memory over each baseline.
  Positive values favor TRACE-Memory; gray shading indicates direction only,
  not statistical significance. Bold indicates $|\operatorname{Adv}|>0.10$.}
  \label{tab:h1-preference}
\end{table*}

\begin{table*}[t]
  \centering
  {\small
    \setlength{\tabcolsep}{2pt}
    \begin{tabular*}{\textwidth}{@{\extracolsep{\fill}}lccccccc@{}}
      \toprule
      & \multicolumn{2}{c}{DeepSeek-V4-Flash}
      & \multicolumn{2}{c}{Gemini 3.1 Flash-Lite}
      & \multicolumn{2}{c}{GPT-5.4} & \\
      \cmidrule(lr){2-3}\cmidrule(lr){4-5}\cmidrule(lr){6-7}
      Model
      & Controlled & Natural
      & Controlled & Natural
      & Controlled & Natural
      & Average \\
      \midrule
      Qwen3-8B
        & \textbf{-0.32} & \textbf{-0.28} & \textbf{-0.21}
        & \textbf{-0.22} & \textbf{-0.38} & \textbf{-0.27} & \textbf{-0.28} \\
      Llama-3.1-8B
        & \textbf{-0.31} & \textbf{-0.25} & \textbf{-0.29}
        & \textbf{-0.23} & \textbf{-0.36} & \textbf{-0.31} & \textbf{-0.29} \\
      Qwen3-14B
        & \textbf{-0.11} & -0.05 & -0.09 & \textbf{-0.13}
        & \textbf{-0.16} & \textbf{-0.11} & \textbf{-0.11} \\
      Qwen3-32B
        & \cellcolor{tracehighlight}0.01 & -0.02
        & \cellcolor{tracehighlight}0.07 & \cellcolor{tracehighlight}0.05
        & -0.09 & -0.03 & 0.00 \\
      \bottomrule
    \end{tabular*}
  }
  \caption{Signed preference advantage of local generators with TRACE-Memory
  over frontier generators with LLM query reformulation and semantic retrieval.
  Positive values favor the local system; bold indicates
  $|\operatorname{Adv}|>0.10$.}
  \label{tab:h2-small-vs-strong}
\end{table*}

\begin{table}[t]
  \centering
  {\small
    \setlength{\tabcolsep}{4pt}
    \begin{tabular}{lcccc}
      \toprule
      Context & \shortstack{EMPTY\\(\%) $\uparrow$}
              & \shortstack{Evidence\\tokens}
              & \shortstack{Anchor\\tokens}
              & Reward \\
      \midrule
      L0 & 7.69  & \textbf{213} & \textbf{2.52} & 0.31 \\
      L1 & 8.03  & 367 & 4.28 & 0.27 \\
      L2 & \underline{10.11} & 294 & 3.40 & \underline{0.41} \\
      L3 & \textbf{12.53} & \underline{257} & \underline{3.25} & \textbf{0.62} \\
      \bottomrule
    \end{tabular}
  }
  \caption{Memory use under public-context interventions. Evidence tokens are
  the realized personal context supplied to the generator; reward is averaged
  over all tasks, including EMPTY decisions. Best and second-best values are
  bold and underlined, respectively; token columns are minimized.}
  \label{tab:h3-policy-statistics}
\end{table}

\begin{table}[t]
  \centering
  {\small
    \setlength{\tabcolsep}{3pt}
    \begin{tabular}{lcccc}
      \toprule
      & \multicolumn{2}{c}{Controlled} & \multicolumn{2}{c}{Natural} \\
      \cmidrule(lr){2-3}\cmidrule(lr){4-5}
      Method & U@20 & U+B@20 & U@30 & U+B@30 \\
      \midrule
      Rule Query             & 57.50  & 58.50 & 51.00  & 55.90 \\
      Oracle Target-Response & \textbf{100.00} & 87.00 & \textbf{100.00} & 87.80 \\
      LLM Query              & \underline{96.20} & \underline{92.70} & 89.90 & \underline{89.10} \\
      TRACE (Qwen3-1.7B)     & 85.70  & 89.80 & 87.50  & 87.00 \\
      TRACE (Qwen3-4B)       & 91.40  & \textbf{96.10} & \underline{91.70} & \textbf{92.80} \\
      \bottomrule
    \end{tabular}
  }
  \caption{Candidate-memory recall (\%). U and B denote useful and borderline
  memories; Controlled and Natural use $K=20$ and $K=30$, respectively. Best
  and second-best values are bold and underlined.}
  \label{tab:h4-candidate-coverage}
\end{table}

\begin{table*}[t]
  \centering
  \includegraphics[width=0.235\textwidth]{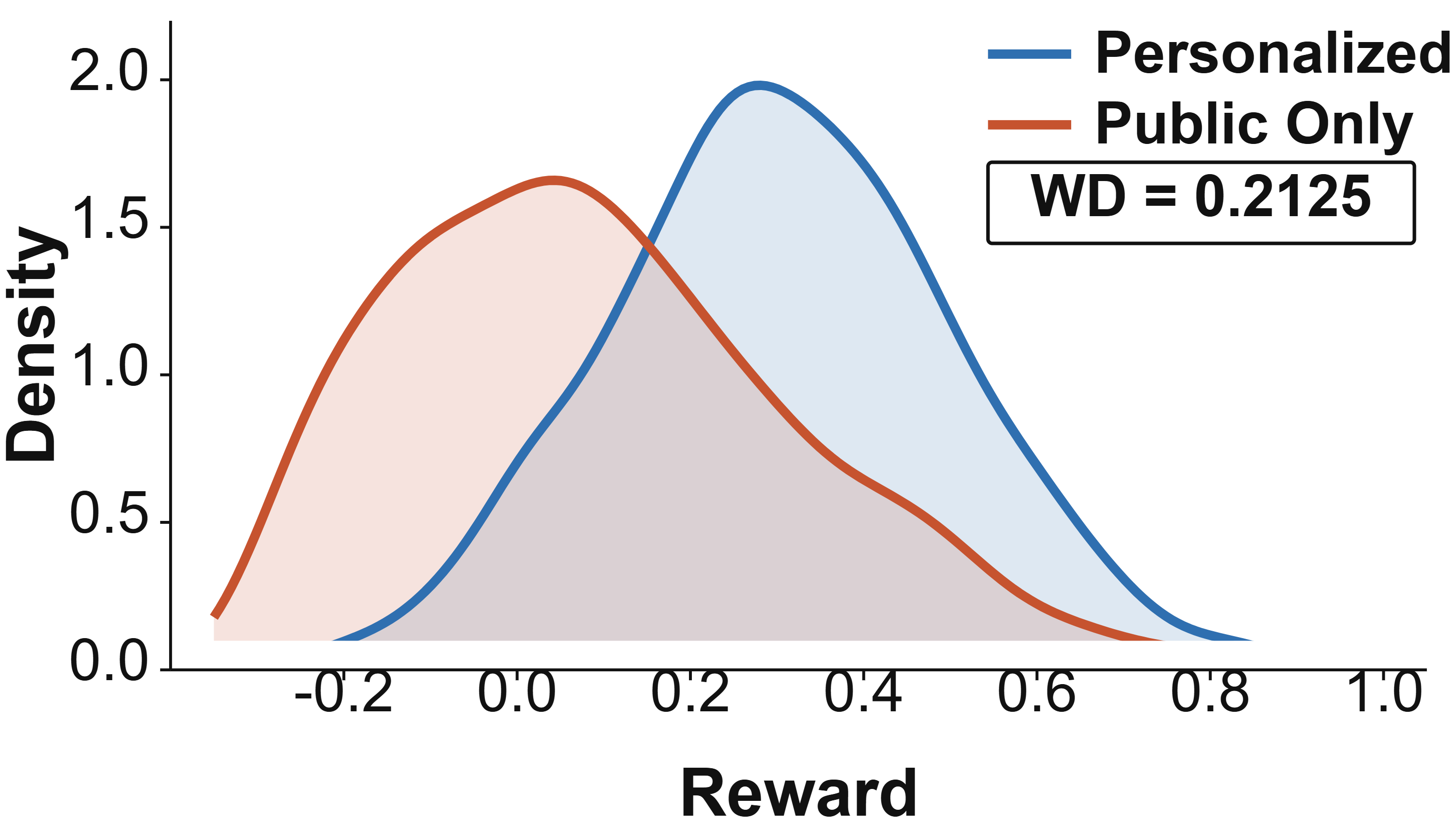}\hfill
  \includegraphics[width=0.235\textwidth]{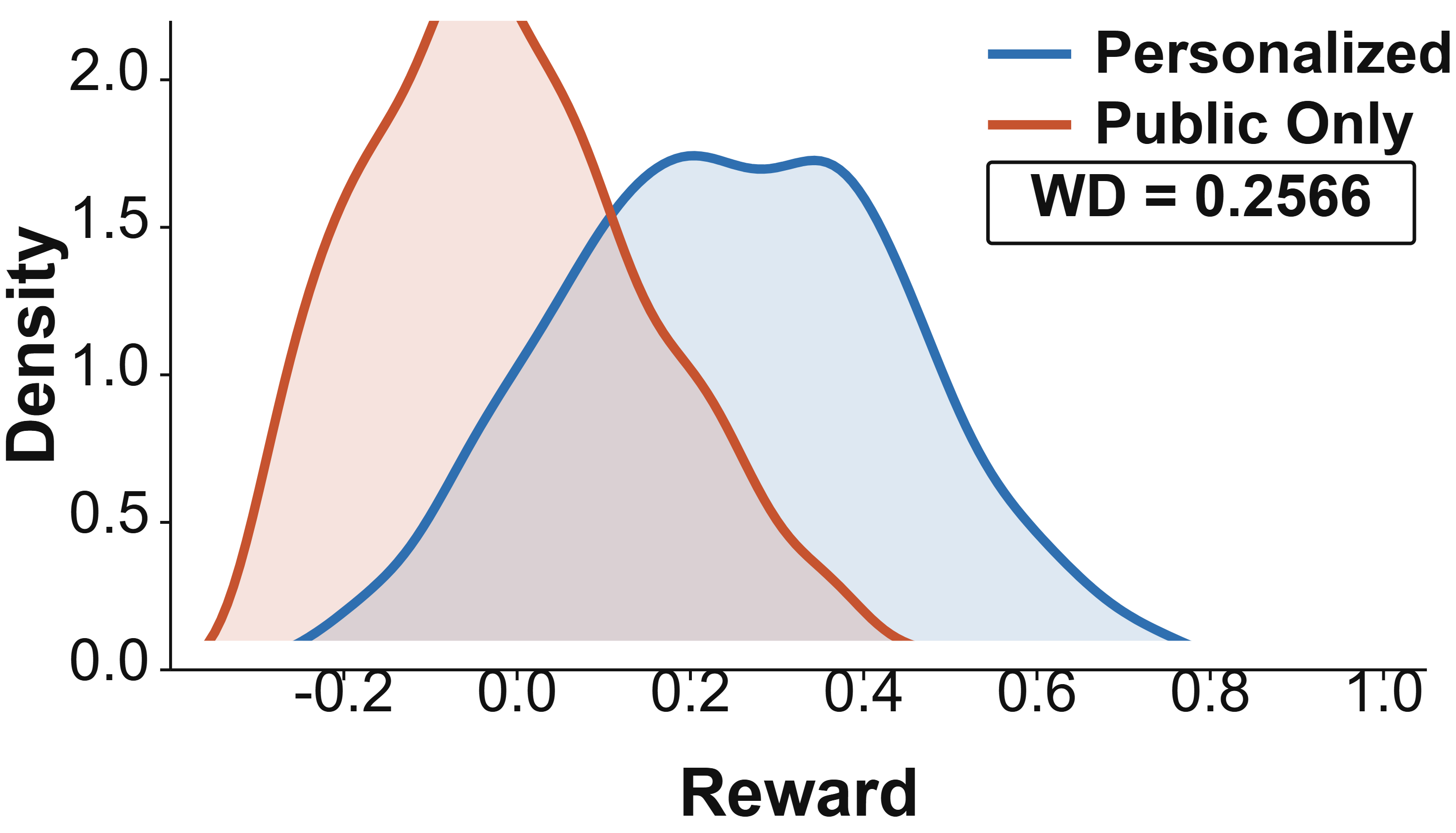}\hfill
  \includegraphics[width=0.235\textwidth]{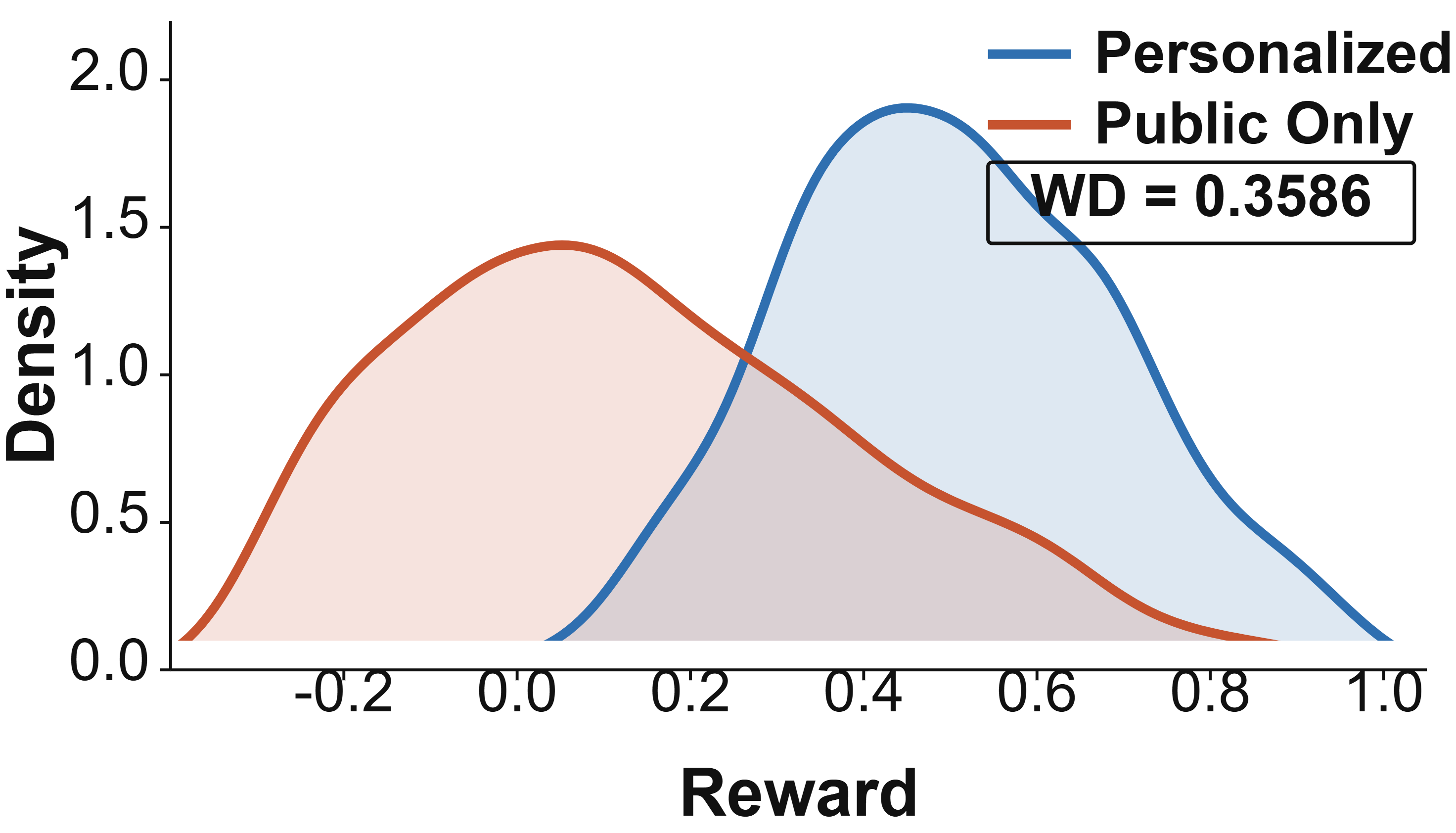}\hfill
  \includegraphics[width=0.235\textwidth]{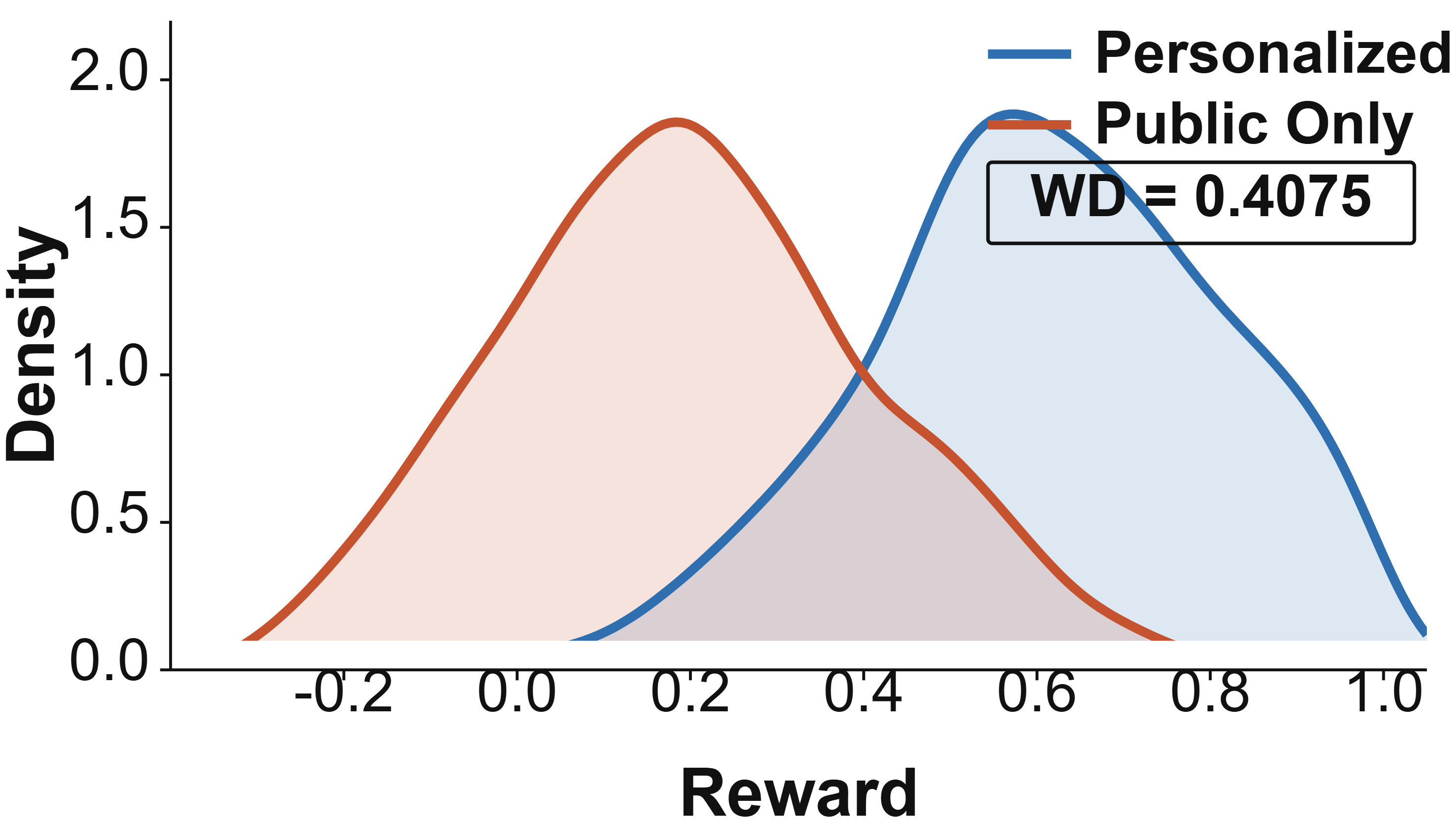}

  \makebox[0.235\textwidth]{\small (a) Qwen3-8B}\hfill
  \makebox[0.235\textwidth]{\small (b) Llama-3.1-8B}\hfill
  \makebox[0.235\textwidth]{\small (c) Qwen3-14B}\hfill
  \makebox[0.235\textwidth]{\small (d) Qwen3-32B}
  {\captionof{figure}{Reward distributions for personalized and public-only
  generation. WD denotes Wasserstein distance.}
  \label{fig:utility-distributions}}

  \vspace{0.8em}
  {\small
    \setlength{\tabcolsep}{2pt}
    \begin{tabular}{llccccccc}
      \toprule
      Dataset & Metric & BM25
      & \shortstack{Semantic\\similarity}
      & \shortstack{LLM query +\\LLM select}
      & \shortstack{Rule query +\\semantic select}
      & \shortstack{Qwen3-4B +\\Qwen3-4B}
      & \shortstack{Qwen3-1.7B +\\Qwen3-4B}
      & \shortstack{Qwen3-4B +\\Llama-3.2-3B} \\
      \midrule
      \multirow{3}{*}{Controlled} & Precision & 3.42  & 9.37  & \underline{60.72} & 26.12 & \textbf{61.09} & 56.83 & 57.93 \\
                                  & Recall    & 17.52 & 23.83 & \textbf{89.22} & 42.47 & \underline{70.12} & 58.52 & 60.17 \\
                                  & F1        & 5.72  & 13.45 & \textbf{72.26} & 32.35 & \underline{65.29} & 57.66 & 59.03 \\
      \midrule
      \multirow{3}{*}{Natural}    & Precision & 3.94  & 17.27 & 32.62 & 39.26 & \underline{48.25} & 47.60 & \textbf{50.29} \\
                                  & Recall    & 21.23 & 27.95 & \textbf{60.24} & 52.78 & \underline{54.48} & 42.14 & 52.30 \\
                                  & F1        & 6.65  & 21.35 & 42.32 & 45.03 & \underline{51.18} & 44.70 & \textbf{51.28} \\
      \bottomrule
    \end{tabular}
  }
  \caption{Candidate-evidence admission precision, recall, and F1 (\%). Model
  pairs are ordered as Stage 1 + Stage 2. Best and second-best values in each
  dataset--metric row are bold and underlined.}
  \label{tab:h4-evidence-admission}
\end{table*}

We evaluate TRACE-Memory through four questions. \textbf{H1} examines
end-to-end quality with the response generator fixed. \textbf{H2} asks whether
better evidence governance helps local generators compete with frontier-model
personalization pipelines. \textbf{H3} tests whether memory admission changes
with the available public context. \textbf{H4} analyzes candidate coverage,
evidence quality, response-level evidence utility, and the contribution of
progressive policy optimization.

\subsection{Experimental Setup}

\noindent\textbf{Data.}
Training uses 4,500 tasks: 2,500 rule-generated Goodreads
\textbf{Controlled} tasks~\citep{wan2018goodreads}, instantiated from predefined
task types and book metadata, and 2,000 \textbf{Natural} tasks from Goodreads,
Amazon Reviews~\citep{hou2024amazonreviews}, and Reddit
\citep{baumgartner2020pushshift} (1,000/500/500). Validation and test each add
450 tasks in the same source proportions, yielding 5,400 total. Each includes a
request,
public context, target-user history, and reference response; target-item
interactions are excluded from memory. Construction and splits are detailed in
the supplementary material.

\noindent\textbf{Models and training.}
Stage 1 uses Qwen3-1.7B or Qwen3-4B~\citep{yang2025qwen3}; Stage 2 uses
Qwen3-4B or Llama-3.2-3B~\citep{meta2024llama32}; both use
LoRA~\citep{hu2022lora}. The BGE-base-en-v1.5 retriever
\citep{xiao2023cpack}, segmenter, and generator remain frozen, while
DeepSeek-V4-Flash~\citep{deepseekai2026deepseekv4} constructs SFT targets.
Training uses structured SFT, reduced-space stage-wise GRPO
\citep{shao2024deepseekmath}, and nested multi-sample Joint GRPO. Local
generators are Qwen3-8B, Llama-3.1-8B~\citep{grattafiori2024llama3}, Qwen3-14B,
and Qwen3-32B; frontier generators are DeepSeek-V4-Flash, Gemini 3.1 Flash-Lite
\citep{google2026gemini31flashlite}, and GPT-5.4~\citep{openai2026gpt54}.
Comparisons share generation settings and evidence budgets; full training
details are in the supplement.

\noindent\textbf{Baselines.}
\textbf{Public-only}, \textbf{Random}, and \textbf{All} use no evidence, random
history, or the full pool. Relevance baselines use \textbf{BM25}
\citep{robertson2009bm25}, BGE-based \textbf{Semantic Similarity}, or
\textbf{Rule-based+Semantic}. \textbf{LLM+LLM} uses a frontier LLM for query
generation and evidence selection: DeepSeek-V4-Flash is fixed for retrieval
analysis, while end-to-end tests use the corresponding frontier generator.
\textbf{Oracle Target-Response Retrieval} uses the unavailable reference only
as a candidate-coverage upper bound.

\noindent\textbf{Evaluation.}
We report Recall@$K$ for useful and useful-plus-borderline memories, and
precision, recall, and F1 for evidence admission. GPT-5.4 and Gemini 3.1 Flash
independently assign useful, borderline, and not useful labels; borderline
disagreements are manually adjudicated. These labelers differ from the
DeepSeek-V4-Flash LLM+LLM baseline.

Following human-preference protocols~\citep{chiang2024chatbotarena}, five
volunteers compare anonymized, order-randomized response pairs under a
double-blind rubric covering task completion, incremental personalization,
evidence faithfulness, personalization appropriateness, and auxiliary language
quality. Majority vote determines the result; ties (under 10\%) are discussed.
The supplementary material provides full annotation details. We report signed
preference advantage:
\begin{equation}
\operatorname{Adv}(A,B)
=
\Pr(A\succ B)-\Pr(B\succ A),
\label{eq:preference-advantage}
\end{equation}
where positive values favor $A$. We also report EMPTY rate, evidence length,
reward mean shift, and Wasserstein distance.

\subsection{End-to-End Personalized Generation Quality}

Table~\ref{tab:h1-preference} tests H1 with the generator fixed. Methods share
inputs, evidence budgets, and generation settings; only memory querying and
selection differ. In LLM+LLM, the column generator also makes memory decisions.
Cells report TRACE-Memory's signed advantage.

TRACE-Memory outperforms Random and BM25 in all six settings, by 0.11--0.38 and
0.05--0.17, respectively. The former exposes the cost of arbitrary history,
while the latter shows the limitation of lexical matching. TRACE-Memory also
beats Semantic Similarity in four settings. Mixed results against Public-only
and All show that neither omitting memory nor injecting the full pool is
reliably optimal; evidence must add value beyond the public path.

LLM+LLM leads in five settings by 0.02--0.12, while TRACE-Memory leads by 0.06
with GPT-5.4 on Natural tasks. That baseline uses the frontier generator for
querying and selection, whereas TRACE-Memory uses trained smaller policies.
Thus, TRACE-Memory reliably improves unfiltered and lexical memory use, usually
improves semantic retrieval, and approaches---but does not uniformly
surpass---frontier-LLM memory processing.

\subsection{System-Level Competitiveness}

H2 compares complete systems: Table~\ref{tab:h2-small-vs-strong} pairs local
generators and TRACE-Memory against frontier generators with LLM query
reformulation and semantic retrieval, testing whether evidence governance can
offset part of the generator gap.

Average signed advantage improves with scale:
$-0.28/-0.29$ for the two 8B generators, $-0.11$ for Qwen3-14B, and
$0.00$ for Qwen3-32B. The 32B system matches or exceeds the
DeepSeek-V4-Flash and Gemini 3.1 Flash-Lite pipelines in three settings but
remains behind GPT-5.4; the 14B deficits narrow to 0.05--0.16. This identifies
a capacity threshold: selective evidence is system-level competitive only with
a sufficiently capable local generator.

Together with H1, this shows that evidence governance offsets part, but not all,
of the gap: GPT-5.4 and both 8B systems retain their respective advantages.
TRACE-Memory's two local policy passes place its complexity between fixed
retrieval and per-instance frontier-LLM processing; without comparable
measurements, we make no hardware-efficiency claim.

\subsection{Public-Context-Aware Selective Memory Use}

H3 fixes the request and history while varying public context.
\textbf{L0} contains metadata, \textbf{L1} same-item review snippets, and
\textbf{L2} aggregated aspect cards. \textbf{L3} is an oracle that uses the
reference response only to select task-relevant public-side information; it
does not place target-user preferences in the public context. L1 and L2 are
alternative representations, so their differences isolate public-context
effects on admission.

Table~\ref{tab:h3-policy-statistics} shows EMPTY rate rising monotonically from
7.69\% at L0 to 12.53\% at L3, especially under structured and oracle public
information. Admission is therefore not fixed by request--history similarity:
the same history is rejected more often as public context covers the task.

L1 yields the longest evidence and lowest reward; L2 and L3 shorten evidence
relative to L1 while improving reward. L0 remains shortest, so length is not
monotonic. H3 is supported at the decision level: public context changes both
memory use and admitted evidence rather than mechanically compressing every
personal context.

\subsection{Internal Mechanism Analysis}

H4 follows candidate coverage, evidence admission, response utility, and policy
optimization.

\noindent\textbf{Candidate coverage and evidence admission.}
Table~\ref{tab:h4-candidate-coverage} evaluates Stage 1 under the
fixed Controlled and Natural budgets ($K=20$ and $K=30$). TRACE with Qwen3-4B
reaches 96.10\% and 92.80\% useful-plus-borderline recall, far above Rule
Query; Oracle Target-Response is an unavailable upper bound.
Table~\ref{tab:h4-evidence-admission} then evaluates the admitted set.
LLM+LLM achieves high recall but only 32.62\% Natural precision, whereas TRACE
variants reach 47.60--50.29\% precision and 44.70--51.28\% F1. Stage 1 thus
preserves candidate coverage while Stage 2 improves evidence purity.

\noindent\textbf{Response-level evidence utility.}
Figure~\ref{fig:utility-distributions} compares personalized and public-only
reward distributions under four local generators. Mean reward shifts are
positive ($+0.21$ to $+0.41$), with Wasserstein distances of 0.2125--0.4075.
The consistent displacement associates admitted evidence with higher
response-level utility across generator scales, although it is not a
controlled context-length ablation. Single-signal reward ablations further
show that using only $\Delta L$ causes policy collapse, whereas using only
$\Delta H$ yields unstable reward trajectories; neither setting completes
training. This motivates combining the two signals in the gated evidence
reward rather than optimizing either one alone.

\noindent\textbf{Progressive policy optimization.}
With Qwen3-4B policies and Qwen3-8B generation,
Table~\ref{tab:h4-training-ablation} compares TRACE-SFT, TRACE-Stagewise, and
TRACE-Full. This comparison directly tests each optimization phase.

\begin{table}[t]
  \centering
  {\small
    \setlength{\tabcolsep}{3pt}
    \begin{tabular}{lccc}
      \toprule
      Variant & \shortstack{Evidence\\F1}
              & \shortstack{Average evidence\\reward}
              & \shortstack{End-to-end\\preference} \\
      \midrule
      TRACE-SFT       & 21.45 & 0.102 & \underline{-0.27} \\
      TRACE-Stagewise & \underline{40.77} & \underline{0.194} & \textbf{-0.15} \\
      TRACE-Full      & \textbf{65.29} & \textbf{0.305} & -- \\
      \bottomrule
    \end{tabular}
  }
  \caption{Ablation over the progressive training curriculum. End-to-end
  preference is the signed advantage of each reduced variant relative to
  TRACE-Full. Best and second-best reported values are bold and underlined.}
  \label{tab:h4-training-ablation}
\end{table}

Evidence F1 rises from 21.45 to 40.77 and 65.29; average reward rises from
0.102 to 0.194 and 0.305. Stage-wise warm-up reduces the TRACE-Full deficit from
$-0.27$ to $-0.15$. Together, coverage, evidence purity, response utility, and
the direct training-stage ablation support H4. The retrieval and admission
comparisons separately diagnose the two stages.

\FloatBarrier

\section{Conclusion}

We proposed TRACE-Memory, which retrieves user-specific information missing
from public context and admits evidence by its incremental response utility,
with an explicit empty-set decision. Across a 5,400-task benchmark, it improves
relevance-based memory use, reaches average parity with frontier pipelines at
32B, and increases abstention as public context becomes sufficient. Mechanism
analysis attributes these gains to candidate coverage, utility-aware admission,
and progressive policy optimization, while stronger generators remain
complementary. Future work will study longer evolving interactions,
cross-generator utility calibration, and matched latency and monetary cost.

\bibliography{references}


\end{document}